\documentclass[11pt]{article}

\usepackage[]{acl}
\usepackage{times}
\usepackage{latexsym}
\usepackage[T1]{fontenc}
\usepackage{longtable}
\usepackage{multirow}
\usepackage{microtype}
\usepackage{inconsolata}
\usepackage{graphicx}
\usepackage{booktabs}
\usepackage{xcolor, colortbl}
\usepackage{pifont}
\usepackage{needspace}
\usepackage[framemethod=TikZ]{mdframed}
\newenvironment{RecBox}[1]
  {\mdfsetup{
    frametitle={\colorbox{white}{\space#1\space}},
    innerleftmargin = 0.25cm, innerrightmargin = 0.25cm, innertopmargin = 0cm, innerbottommargin = 0.25cm,
    frametitleaboveskip=-\ht\strutbox,
    frametitlealignment={\hspace{-5pt}},
    nobreak=true,
    skipabove=10pt,
    skipbelow=7pt
    }
  \begin{mdframed}%

  }
  {\end{mdframed}}

\newenvironment{LessonBox}
  {\mdfsetup{topline=false,bottomline=false,rightline=false,
    linewidth=2pt,linecolor=black!30,
    innerleftmargin=8pt,innerrightmargin=0pt,innertopmargin=2pt,innerbottommargin=2pt,
    skipabove=6pt,skipbelow=6pt}
  \begin{mdframed}\small\noindent\textbf{Lesson}\hspace{0.6em}}
  {\end{mdframed}}

\newcommand\blfootnote[1]{%
  \begingroup
  \renewcommand\thefootnote{}\footnote{#1}%
  \addtocounter{footnote}{-1}%
  \endgroup
}
\title{Bye Bye \textsc{perspective api}: \\Lessons for Building and Governing Measurement Infrastructure}

\author{
  \textbf{David Hartmann\textsuperscript{1,2}},
  \textbf{Manuel Tonneau\textsuperscript{1,3}},
  \textbf{Angelie Kraft\textsuperscript{1}},
  \textbf{LK Seiling\textsuperscript{1}},
  \textbf{Dimitri Staufer\textsuperscript{2}},
  \\
  \textbf{Pieter Delobelle\textsuperscript{4,5}},
  \textbf{Jan Fillies\textsuperscript{6}},
  \textbf{Anna Ricarda Luther\textsuperscript{7,8}},
  \textbf{Jan Batzner\textsuperscript{1,9,10,\dag}},
  \textbf{Mareike Lisker\textsuperscript{11,12,\dag}}
  \\
  \\
  \textsuperscript{1}Weizenbaum Institute,
  \textsuperscript{2}TU Berlin,
  \textsuperscript{3}University of Oxford,
  \textsuperscript{4}KU Leuven,
  \textsuperscript{5}Pleias,
  \\
  \textsuperscript{6}Freie Universität Berlin,
  \textsuperscript{7}University of Bremen,
  \textsuperscript{8}ifib research,
  \textsuperscript{9}TU Munich,
  \\
  \textsuperscript{10}Munich Center for Machine Learning,
  \textsuperscript{11}HTW Berlin,
  \textsuperscript{12}University of Hamburg
  \\
  \small{
    \textbf{Correspondence:} \href{mailto:d.hartmann@tu-berlin.de}{d.hartmann@tu-berlin.de}
  }
}

\begin{document}
\maketitle
\blfootnote{\textsuperscript{\dag}These authors contributed equally.}
\begin{figure*}[t!]
    \centering
    \includegraphics[width=0.86\linewidth]{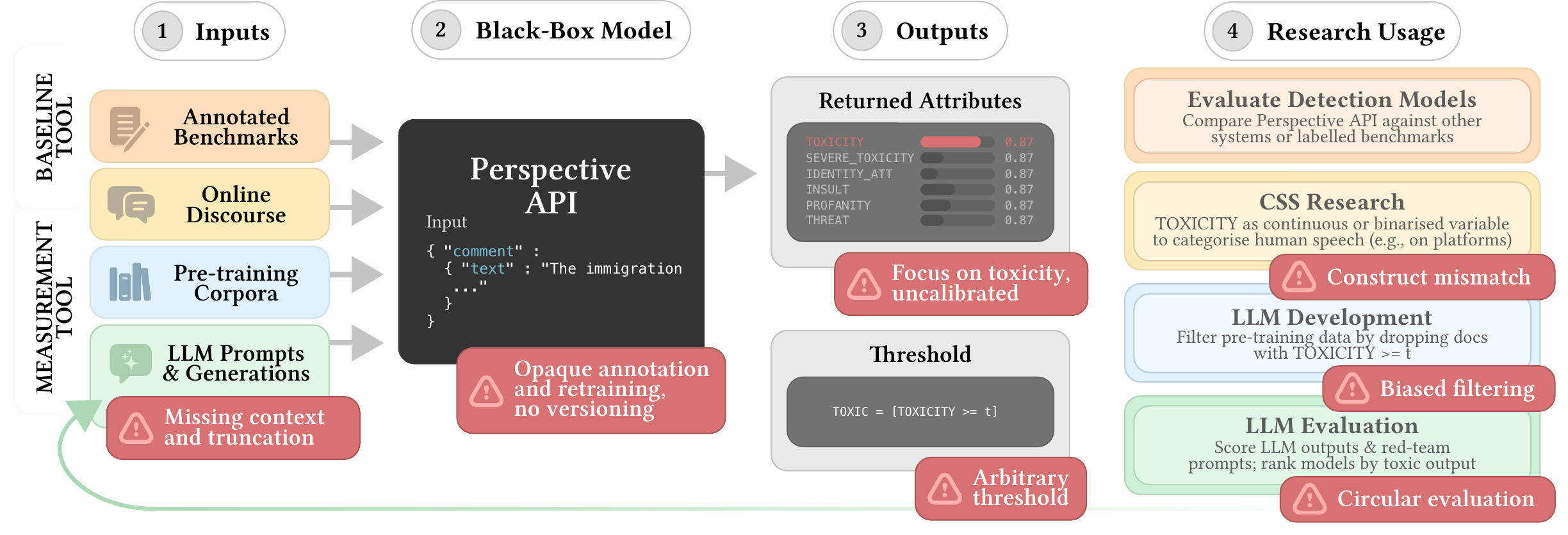}
     \caption{\textbf{How \textsc{perspective api} is used in research, as a \textit{subject of} evaluation and as a \textit{measurement tool}.} The figure shows choices in operationalisation and provision that are critiqued in this paper: \ding{192}~\textbf{Inputs} are scored as isolated sentences, stripping sender, target, and conversational context. \ding{193}~The \textbf{Black-Box Model} is opaque about model updates and annotations. \ding{194}~\textbf{Outputs} include six attributes, but downstream work retains only \textsc{Toxicity} and binarises its uncalibrated score at an ad hoc threshold. \ding{195}~\textbf{Research Usage} covers evaluating detection models, CSS research, LLM development, and LLM evaluation, with issues such as construct mismatch, biased filtering, and circular evaluation. \S\ref{sec:critique} and \S\ref{sec:struct_circ} develop each point and its consequences for research quality.}
    \label{fig:main}
\end{figure*}

\begin{abstract}
\textsc{perspective api} closes at the end of 2026, removing the de facto standard for toxicity measurement and exposing researchers' dependence on a tool they did not control. Drawing on this case, we argue that a research field must build and govern its own measurement infrastructure rather than borrow it. Surveying 241 papers that use or study \textsc{perspective}, we show what depending on it cost the research community: claims reaching past what the tool could support, results that shifted when its model was silently retrained, and disparities researchers could measure but not explain. These failures were amplified throughout the LLM lifecycle, where \textsc{perspective} supplied the labels, filtered the corpora, and graded the systems trained on each, rewarding errors rather than catching them. To keep the instrument open to study after shutdown, we release \textsc{perspective} scores for 5.9 million text snippets from 77 datasets. We further specify ten requirements for measurement infrastructure a field owns, and argue that what blocks such infrastructure is not technical capability but the value the field places on infrastructure work. 
\end{abstract}

\section{Introduction}
Online abuse can have profound real-world consequences, including the suppression of marginalised voices, deterioration of mental health, and violence against marginalised groups \citep{marques2023, Luther2025}. Research responded with automated classifiers that detect such content at scale \cite{vidgen2020directions}. One became particularly influential: \textsc{perspective api}, developed by Google Jigsaw. Since 2017, it has offered a free, scalable API for classifying \textit{toxicity}, an umbrella term for harmful content, and is now widely described as ``a foundation for academic research on online abuse and incivility'' \citep{nogara2025}.

Despite its popularity, Jigsaw recently announced that \textsc{perspective api} will cease operation at the end of 2026, because, in its words, ``As AI capabilities evolve, we are setting our sights on new challenges'' \citep{jigsaw2026}. For communities that built their work around it, the shutdown suddenly withdraws critical infrastructure, exposing how far the field had come to depend on a proprietary, externally controlled measurement tool.

Against that dependence, this position paper argues from the example of \textsc{perspective api} that a field must build and govern the measurement infrastructure it relies on. We analyse a decade of \textsc{perspective api} and what relying on it cost the research community: findings that outran what the tool was validated for, results that could not be compared over time, and biases that could not be explained. From this, we draw lessons for measurement infrastructure a field owns, from how its construct is defined to how the work is funded.

We make four contributions. \textit{First}, we present a systematic bibliographic review of 241 papers that use or study \textsc{perspective}, tracing how it was used across hate speech detection, computational social science (CSS), and LLM development and evaluation, and what drove its adoption (\S\ref{sec:what}). \textit{Second}, judged against standard procedures of measurement validation, we show \textsc{perspective} was never fit for the central role it came to play (\S\ref{sec:critique}), and that these failures compound when the same tool filters pretraining data and grades the models built from it (\S\ref{sec:struct_circ}). \textit{Third}, from those failures, we derive five technical and five governance requirements (T1--T5, G1--G5), most of which apply to any measurement infrastructure, and assess existing alternatives against them (\S\ref{sec:infra}). \textit{Fourth}, we release \textsc{perspective} scores for 77 datasets covering 5.9 million text snippets, keeping the instrument open to study after shutdown (\S\ref{sec:infra}). We close by asking why no alternative was built, and what would have to change for the field to build one (\S\ref{sec:concluding}).

\section{\textsc{perspective api} and Its Research Uptake}
\label{sec:what}
\subsection{Use in Research}
Within academic research, \textsc{perspective} was taken up across communities in two distinct ways (Figure \ref{fig:main}): as a \textit{subject of} evaluation and as a \textit{measurement tool}. To measure this uptake, we conducted a systematic bibliographic search of research using \textsc{perspective} (see Appendix~\ref{app:bibliography} for details). We found 241 papers, 18 of which are meta-critiques, and labelled the remaining 223 by usage type: baseline model, CSS measurement, pre-training filtering, or LLM evaluation.

\paragraph{\textsc{perspective} as a baseline model} 
In \textit{toxicity} and \textit{hate speech detection} research, \textsc{perspective} was primarily treated as a system to be evaluated ($n = 83$). \citet{pavlopoulos-etal-2019-convai}, for example, assess \textsc{perspective} as a baseline for the SemEval-2019 offensive language task. Evaluations of this kind measure agreement with a benchmark's labels. Where the benchmark encodes a different construct, such as offensive language or hate speech, the result shows how closely \textsc{perspective} tracks that construct, not whether toxicity was measured well. Where the benchmark encodes toxicity, the labels usually descend from the same annotation scheme that produced \textsc{perspective}'s training data, so agreement demonstrates fidelity to that scheme rather than support for the construct the scheme encodes. Neither result establishes the validity of Jigsaw's toxicity construct.

\paragraph{\textsc{perspective} as a measurement tool}
In contrast, CSS  and LLM development and evaluation research primarily used \textsc{perspective api} as a tool to measure toxicity at scale. In CSS, researchers applied the API in large-scale studies of online discourse ($n = 50$, e.g., \citealp{theocharis2020, kim2021, chang2023, frimer2023, mosleh2024misinfo, gervais2025}). In LLM development, \textsc{perspective} was used to filter pre-training data by removing text flagged as toxic from the corpora used to train and align LLMs ($n = 29$, e.g., \citealp{welbl2021}). In LLM evaluation (benchmarking, auditing, and red teaming), \textsc{perspective} was built into evaluation operationalisations to label model outputs and, in red teaming, to flag potentially offensive inputs ($n = 61$, e.g., \citealp{gehman2020, liu2021dexperts, lee2023query, liang2023}). Across these three uses, \textsc{perspective} is not the object of study but the instrument that produces the finding. Whatever the tool captures or misses therefore passes into the published result, whether that is a claim about discourse or a safety score. 

\subsection{Drivers of Adoption}\label{sec:adoption}

At its peak, \textsc{perspective} processed 500 million requests daily \citep{jigsaw2021perspective}, and was integrated into comment systems at major publishers, including the New York Times, Vox Media, and OpenWeb \citep{jigsaw2021perspective, rieder2021}, while also becoming widely adopted in academic research \citep{ribeiro2024}. In what follows, we examine the factors behind this widespread adoption. 

\paragraph{Timing} \textsc{perspective}'s launch in 2017 followed a surge in online abuse, including high-profile harassment campaigns \citep{massanari2017}, the spread of conspiratorial narratives \citep{venturini2019junk}, inter-ethnic violence and genocide \citep{mozur2018genocide}, and the mobilisation of political extremism online \citep{ganesh2020countering}. Regulators responded: Germany’s \textit{Network Enforcement Act} (NetzDG), passed in 2017, required platforms to remove clearly illegal content. This moved the burden of moderation from users to platforms, which turned to centralised and increasingly automated systems \citep{gillespie2018custodians, gorwa2019, gorwa2020}. It also motivated academic research on automatic hate speech detection \citep{waseem-hovy-2016-hateful}. \textsc{perspective} thus emerged at the intersection of regulatory pressure and a nascent research field, offering scale where human moderation could not \cite{davidson2017, gorwa2020}.

\paragraph{Cost and access} \textsc{perspective api} was free throughout its decade of operation, whereas competing tools such as Amazon's \textsc{rekognition} charge per image analysed \citep{aws_rekognition_pricing}. Access was provided upon application to Jigsaw, though acceptance criteria remained opaque \cite{rieder2021}. The API offered a 100ms response time and an expandable quota on request, making large-scale studies feasible at no cost. Its coverage of 18 languages, trained on 38 million records \citep{lees2022}, further enabled research beyond English \citep{tonneau-etal-2024-languages}. 

\paragraph{Ease of use, transparency, and performance} \textsc{perspective} accepted single-sentence inputs without preprocessing and returned an interpretable scalar score, lowering the barrier to integration in research and production alike. Jigsaw provided model cards, threshold guidance, and partial architectural documentation \citep{perspectiveapimodelcard}, a level of transparency uncommon among commercial APIs \citep{hartmann2025}. The team also engaged the research community through publications, benchmark datasets \citep{dixon2018, lees2022}, and responses to researcher feedback \citep{perspectiveapi_research}. Though incomplete, as the next section details, this documentation positioned \textsc{perspective api} as a comparatively transparent and research-embedded proprietary tool. \textsc{perspective} also outperformed academic hate speech models in real-world evaluations, despite performing poorly in absolute terms \citep{tonneau-etal-2025-hateday}.

\paragraph{Institutional credibility} Adoption was also driven by institutional backing. \textsc{perspective} came from Jigsaw, a Google subsidiary, and grew out of a collaboration with Wikimedia \citep{wulczyn2017}, two institutions whose standing lent credibility to the tool. Jigsaw also supplied the corpora and public competitions the field then built on \citep{jigsaw_kaggle}. External validation followed: the New York Times, itself a recognised institution, adopted a moderation tool built with Jigsaw \citep{nyt2018perspective}. This institutional weight is what made \textsc{perspective} credible before it was validated, and it is likely why adoption continued despite the biases documented in \S\ref{sec:critique}.

\paragraph{Path dependence and network effects}
As \citet{star1999ethnography} argues, tools become infrastructure through their embeddedness in practice. Once taken up across a field, a measurement instrument shapes what is measurable, comparable, and credible, generating path dependence \citep{bowker1999}. \textsc{perspective api} exemplifies this dynamic: its integration into benchmarking suites such as \textit{RealToxicityPrompts} \citep{gehman2020} and the \textit{Holistic Evaluation of Language Models} (HELM) \citep{liang2023}, as well as its use in CSS studies of political incivility and online harassment \citep{theocharis2020, gervais2025}, positioned it as a common reference point across research contexts. This standardisation enabled comparability across studies, which in turn encouraged continued use, creating a self-reinforcing dynamic consistent with network effects \citep{katz_systems_1994}. In this sense, \textsc{perspective api} was not necessarily adopted because it was the most valid or best-performing tool, but because it had become an established point of reference, effectively positioning it over time as the de facto standard for toxicity detection. 

\section{Was \textsc{perspective api} Ever Fit for Purpose? A Measure Development and Infrastructural Critique} \label{sec:critique}

Becoming a standard through path dependence and network effects is not the same as being validated as one. In this section, we assess whether \textsc{perspective api} was fit for the central role it came to play in research by comparing its development to established procedures of measurement validation in the social sciences \citep{clark2019, jacobs, wallach2024}. 

In this tradition, a \textit{concept} (i.e., a broad and often contested notion) is first systematised into a \textit{construct} (i.e., a more specific and explicitly defined understanding), specifying \textit{what} is to be measured \citep{adcock2001validity}. The construct is then operationalised through a \textit{measurement instrument}, specifying \textit{how} it is to be measured. Both steps, \textbf{systematisation} and \textbf{operationalisation}, are inherently situated and involve analytical choices \citep{wallach2024}. They rely on assumptions that must be made explicit and supported theoretically and empirically.\footnote{This is commonly referred to as \textit{construct validation}, the process of establishing whether a measurement instrument captures what it is intended to measure or, in other words, of providing theoretical and evidential support for one's operationalisation \citep{cronbach1955construct}.}

We show that \textsc{perspective} provides limited grounding for key assumptions in its systematisation (\S\ref{subsec:systematisation}) and operationalisation (\S\ref{subsec:operationalisation}), and that these limitations affect performance and research use (\S\ref{subsec:research_quality}). We further identify structural issues in its application, particularly in cases where it is used to both define and measure toxicity (\S\ref{sec:struct_circ}).

\subsection{Systematisation}\label{subsec:systematisation}

\paragraph{Contested behavioural assumption}
\textsc{perspective}'s construct of toxicity rests on a behavioural assumption that is empirically contested. Jigsaw defines toxicity as ``a rude, disrespectful, or unreasonable comment that is likely to make people leave a discussion'' \citep{dixon2018, rieder2021}, thereby tying toxicity to disengagement. However, empirical evidence suggests that this effect is not stable. For example, \citet{beknazar2025toxic} show that removing toxic content from social media feeds causally reduces time spent, ad impressions, and content consumption, indicating that its presence contributes to user engagement. Observationally, out-group animosity is the strongest predictor of shares and retweets on Facebook and Twitter \citep{rathje2021out}, and toxic conversations on Twitter tend to generate larger, wider, and deeper reply trees than non-toxic ones \citep{saveski2021structure}. Taken together, this evidence challenges the disengagement premise: the same toxic content can repel some readers and engage others, depending on their relationship to it. A construct that ties toxicity to a single, universal behavioural effect therefore mistakes a reader-dependent phenomenon for a property of the text.

\paragraph{Insufficiently specified construct}
Even beyond this contested behavioural assumption, the construct remains insufficiently specified. \textsc{perspective api} provides scores for multiple ``toxic attributes'', including threat, insult, identity attack, severe toxicity, and profanity, suggesting a decomposition of toxicity into subcomponents. However, Jigsaw gives only one-sentence definitions and does not distinguish between the attributes, relate them to the overarching construct, or say whether the set is exhaustive \citep{rieder2021, hartmann2025}. Both the scope of the construct and its relationship to these subcomponents therefore remain unclear, making it difficult to determine what is being measured or to assess construct validity \citep{wallach2024}.

\paragraph{Construct-measure mismatch}
This lack of specification enables a mismatch between the intended construct and its use in downstream research, particularly in studies of hate speech. Toxicity and hate speech are related but distinct \citep{barth2023}: toxicity covers rude or unreasonable language generally, while hate speech targets individuals or groups on protected characteristics \citep{anderson2023, tonneau-etal-2024-languages}. A measurement instrument designed for toxicity therefore cannot be assumed to validly capture hate speech, a conflation common across the field's datasets and literature \citep{vidgen2020directions}. In this regard, we observe three recurring usage patterns in the literature. The first uses \textsc{perspective}'s toxicity score to measure toxic language, aligning with the tool's intended use, though constrained by its behavioural definition. The second uses attributes such as \texttt{IDENTITY\_ATTACK} as proxies for hate speech, typically acknowledging the construct gap explicitly (e.g., \citealp{tonneau-etal-2025-hateday}). The third frames research questions in terms of hate speech, political incivility, or LLM safety and operationalises them using the toxicity score without acknowledging the lack of theoretical alignment \citep{kim2021, frimer2023}. This last pattern is the most problematic: the instrument does not match the target construct and the mismatch goes unstated, a practice \citet{gervais2025} document for political incivility research. 

\begin{LessonBox}
\textsc{perspective} tied toxicity to a contested behavioural effect and never fixed the scope of its attributes, leaving a construct too vague to validate and loose enough to absorb hate speech, incivility, and LLM safety alike. Any successor should state its construct explicitly and say for whom it holds (T2, G3).
\end{LessonBox}

\subsection{Operationalisation and Provision}\label{subsec:operationalisation}

We now turn to how toxicity is operationalised and provisioned in \textsc{perspective api}. We focus on three aspects: the input representation, the model and its provision, and the resulting outputs.

\paragraph{Input: context-free}
\textsc{perspective api} takes a single sentence or short text as input and returns a toxicity score, thereby operationalising toxicity as a property identifiable in isolated text \citep{rieder2021, hartmann2025}, as most classifiers of the period did. However, the meaning of a sentence depends on its conversational context, community norms, who is speaking, who is targeted, and the relationship between speakers and receivers \citep{mosca2021, yoder2022, anderson2023}. By design, \textsc{perspective api} strips this away, limiting the instrument to surface features and excluding the relational dimensions that define toxicity as a social phenomenon. The same text therefore receives the same score across contexts, even though its perceived toxicity depends on context-specific norms.

\paragraph{Model: misaligned and opaque}
The annotation scheme, while partially available \citep{conversationai_annotation}, diverges from Jigsaw’s public definition of toxicity \citep{dixon2018, rieder2021}. Whereas the public definition frames toxicity as a conjunctive criterion, namely a rude or disrespectful comment that is likely to make users leave a discussion, the annotation scheme applies these criteria disjunctively, allowing labels to be assigned based on rudeness or anticipated behavioural effects alone. As a result, the model is trained on labels that do not correspond to the definition users are given, creating a gap between construct and operationalisation. Opacity compounds the mismatch at the level of data, model, and input. The annotation process went undocumented, including annotator demographics, data sources, and labelling procedures \citep{rieder2021, hartmann2025}. The model was silently retrained, an update warranted by evolving slurs and coded language \citep{siegel2020online, mendelsohn-etal-2023-dogwhistles} but made without version pinning or changelogs, and announced only in a group Jigsaw said could not cover every update \citep{pozzobon-etal-2023-challenges}. Finally, the system scored only the first 501 characters of each input, a limitation the published documentation omits \citep{talas2024perspective}.

\paragraph{Output: uncalibrated}
The outputs of \textsc{perspective api} are not calibrated to any interpretable real-world quantity. The API returns a score between 0 and 1, but this score does not correspond to prevalence, severity, or annotator agreement, as is true of scalar classifiers generally \citep{guo2017calibration}. According to Jigsaw’s technical documentation, scores are normalised against a balanced 50/50 toxic and non-toxic dataset rather than any real-world distribution \citep{jigsaw2017normalization}, meaning they reflect model confidence under an assumed even split \citep{nogara2025} rather than the prevalence of harm. Threshold guidance existed \citep{hua2020characterizingtwitterusersengage, hua2020measuringadversarialtwitterinteractions}, but those thresholds were validated for particular corpora and tasks, and in practice, the choice remained an undocumented decision made study by study. As a result, scores lack a stable interpretation and depend on arbitrary and often unreported thresholding choices.

\begin{LessonBox}
\textsc{perspective} stripped context from its inputs, trained on labels that diverged from its published definition, retrained silently, and returned a score normalised to a distribution no real corpus resembles. Any successor should document and version every such design decision, and accept the context its construct depends on (T1, T3, T4, G2).
\end{LessonBox}

\subsection{Consequences for Research Quality}\label{subsec:research_quality}
As \textsc{perspective api} became embedded in research workflows, the limitations above produced systematic errors, inconsistent results, and biases across use cases.

\paragraph{Lacking validity and performance}
Construct validity issues translate into poor classification performance when \textsc{perspective} is applied beyond its intended scope. \citet{gervais2025} show that \textsc{perspective} scores on congressional tweets achieve a maximum F1 of 0.33 against human-coded political incivility across all decision thresholds, and \citet{mihaljevic2022} find that it detects antisemitism only at a very basic level. Failures are most pronounced where interpretation depends on context: counterspeech is labelled hateful \citep{gligoric-etal-2024-nlp}, reclaimed slurs are over-moderated \citep{diasoliva2021, rottger2021}, and implicit hate speech, which relies on background knowledge, is systematically missed \citep{hartmann2025}. Finally, in real-world settings, \textsc{perspective} performs very poorly for hate speech detection but still outperforms academic models, pointing to a broader model miscalibration in the field \citep{tonneau-etal-2025-hateday}.

\paragraph{Undermined reliability and reproducibility}
Silent retraining made results from different times incomparable.
\citet{pozzobon-etal-2023-challenges} show that rescoring \textit{RealToxicityPrompts} with a later version of the API moved around 10K prompts from toxic to non-toxic and 232 in the opposite direction, a 49\% net reduction in prompts labelled toxic. The same rescoring produced a different ranking of the 37 models in the HELM toxicity benchmark. The shutdown makes this problem permanent: scores can no longer be recovered or verified, rendering \textsc{perspective api}-based benchmarks irreproducible.

\paragraph{Systematic bias and limited attribution}
Bias is evident in \textsc{perspective}'s outputs, but its sources cannot be attributed, given limited transparency about the data and annotation. In Jigsaw's own early evaluations, benign sentences such as \textit{``I am a gay man''} scored as toxic purely for naming an identity \citep{dixon2018}. That Jigsaw published this failure and then ran a public competition on it \citep{jigsaw_kaggle} is rare among commercial providers. Later work broadened this: LGBTQIA+ content and reclaimed slurs are disproportionately flagged \citep{diasoliva2021, rottger2021}, as is African-American English, a pattern documented for academic classifiers as well \citep{sap-etal-2019-risk, xu-etal-2021-detoxifying}, while hate against marginalised groups is underdetected \citep{hartmann2025}. These patterns suggest reliance on surface features such as identity terms and dialect markers, combined with context-free inputs. Biases extend across languages and geographies: German text is rated more toxic than comparable content elsewhere \citep{nogara2025}, and \textsc{perspective} scores an abusive Hindi word at 0.10 while the same insult in English scores 0.97 \citep{ghosh-etal-2021-detecting}. Annotator demographics, beliefs, and guidelines shape what is labelled toxic in edge cases \citep{sap2022, davani2023}, and because \textsc{perspective} documents none of them (\S\ref{subsec:operationalisation}), the disparities above are measurable but not explicable: whether dialect is over-flagged because of who annotated, what they were told to do, or which text they saw cannot be established from outside.

\begin{LessonBox}
\textsc{perspective}'s users published findings beyond its validated scope, could not compare results across time, and could not attribute the biases they observed. Any successor should validate performance within a stated scope, tie every result to a version, and document annotation well enough that disparities can be traced (T1, T2, G2).
\end{LessonBox}

\section{Amplification and Circularity across the LLM Lifecycle}\label{sec:struct_circ}

These issues extend beyond individual studies. Because \textsc{perspective api} is embedded at both ends of the LLM lifecycle, its errors propagate into the behaviour and standards of widely deployed LLMs.

\paragraph{Biased data filter} A common step in LLM pretraining (\S\ref{sec:what}) is to score the corpus with \textsc{perspective} and drop what it flags. Google and DeepMind report doing this \citep{welbl2021, longpre2024pretrainersguide}, and Cohere a related pipeline \citep{ngo2021mitigating}. The filtering happens before any model exists, so whatever the classifier misjudges is simply absent from what the model can learn. The downstream cost is not borne equally. As shown in \S\ref{subsec:research_quality}, \textsc{perspective} systematically over-flags African-American English, LGBTQIA+ content, and reclaimed slurs. Filtering with it therefore removes these categories disproportionately from training corpora \citep{xu-etal-2021-detoxifying}, leaving models worse at understanding these communities and at producing the speech and values their language carries. This compounds \textsc{perspective}'s silencing effect: communities whose speech the classifier already over-moderates may now also be under-served by the LLMs trained on its filtered output. The same tool then scores those models, so the check cannot see what the filter removed.

\paragraph{Circular evaluation} Beyond filtering, \textsc{perspective api} also generated the training labels and served as the evaluation standard for classifiers built on them, producing a self-reinforcing circularity that runs through the field's most influential resources, including \textit{RealToxicityPrompts} \citep{gehman2020, pozzobon-etal-2023-challenges}. The loop has three recurring steps. \textit{First}, researchers used \textsc{perspective} scores to generate training labels for toxicity and hate speech datasets, and for controllable-generation methods such as DExperts \citep{gehman2020, liu2021dexperts}. \textit{Second}, models trained on these labels were evaluated against \textsc{perspective} outputs, directly or through benchmarks it had labelled \citep{gehman2020, liang2023, pozzobon-etal-2023-challenges}. \textit{Third}, \textsc{perspective} then graded the models built from its own labels \citep{liu2021dexperts, zia2022toxicity}. At each step, ad hoc thresholds such as $t=0.7$, often unreported (see stage~\ding{194} in Figure~\ref{fig:main}), turned uncalibrated scores into binary ground truth \citep{pozzobon-etal-2023-challenges, hartmann2025}.

Across both filtering and evaluation, the field's working definition of toxicity collapsed into a single system's decision threshold. Because the same tool shapes both what models learn and what they are graded on, its systematic errors become invisible at both ends: models that inherit \textsc{perspective}'s biases score well on \textsc{perspective}-derived benchmarks because the errors are in the standard itself. The result is amplification rather than correction, since an error introduced when the corpus is filtered is confirmed when the model is graded. This is what \citet{blodgett2020} describe as undeclared social assumptions laundered through technical framing until they appear neutral. Once \textsc{perspective} became the measure, it ceased to function as one \citep{Strathern_1997}.

\begin{LessonBox}
When one tool filters the training data and grades the models built from it, the field's working definition of toxicity collapses into that tool's decision threshold, and its errors become invisible at both ends. An instrument must be independent of what it evaluates (G1).
\end{LessonBox}

\section{What Now? Towards Independent Research Infrastructure}
\label{sec:infra}

\textsc{perspective api} emerged when platforms were investing in moderation under regulatory pressure (\S\ref{sec:adoption}). That investment has since reversed. Since 2024, platforms have reduced moderation workforces \citep{rod_mcguirk_x_2024, gerken_tiktok_2025, mcmahon_facebook_2025}, loosened content rules \citep{biddle2025meta}, curtailed researcher data access \citep{burnat2025accountability, hartmann2025addressing}, and dismantled trust and safety teams \citep{moran2025endts}, withdrawing from earlier moderation commitments \citep{magalhaes2026} even as toxic speech rises \citep{hickey2025x, tornberg2025elite} and remains implicated in offline violence \citep{cnn2025ballymena, euronews2025torrepacheco, venkataramakrishnan2025disinformation}. Regulation has expanded over the same period. Platforms in the EU now face the Digital Services Act \citep{dsa_2022} and the AI Act \citep{eu_ai_act_2024}, and in the UK the Online Safety Act \citep{uk_osa_2023}. Each imposes duties that presuppose harmful content can be measured, since enforcement and compliance alike depend on it. Independent measurement capacity therefore matters more now than ever. We set out what the community should do before the shutdown, and the requirements that a successor must meet.

\paragraph{Immediate actions} Before the shutdown on 31 December 2026, the community should archive \textsc{perspective} scores for all datasets and benchmarks, including full method documentation such as query date, threshold settings, and API version when available. These should be registered as versioned data artefacts. We contribute to this effort by releasing a snapshot of 77 publicly available toxicity datasets and major benchmarks, with 5.9 million scored text snippets \cite{hartmann2026snapshot}. The release keeps \textsc{perspective} open to study once the API is gone, and gives successor systems a fixed public baseline to compare against (Appendix~\ref{app:snapshot}). These are short-term measures, not the successor itself. Until one exists, new projects should use open-weight models with clear versioning instead of closed endpoints, and any LLM used as a judge for toxicity detection should have its parameters and system prompts disclosed. The snapshot should serve only as a fixed reference for historical comparison. New evaluations should use human labels where possible, or calibrated open-weight classifiers reported alongside them.

\paragraph{Requirements and recommendations} From the limitations in \S\ref{sec:critique}, we derive ten requirements for a successor infrastructure, grouped into \textbf{T}echnical and \textbf{G}overnance dimensions and mapped in Table~\ref{tab:requirements} to the failures they address. Seven of the ten apply to any measurement infrastructure. Uncertainty (T4) and legitimacy (G3) also assume a construct that people disagree about, and only contextuality (T3) is specific to abusive language.

\begin{RecBox}{T1 \textperiodcentered\ Reproducibility}
Record the model, dataset, and metric behind every evaluation, and pin a citable version. \end{RecBox}
\vspace*{-0.5em}
\noindent Versioning in a community-agreed schema \citep{biderman2024lessons,batzner2026everyevalever} is what makes a repeated query a repeated measurement, and what lets a result be rechecked years later.

\begin{RecBox}{T2 \textperiodcentered\ Validity}
State which construct the tool measures, under what conditions, and where that construct stops applying. \end{RecBox}
\vspace*{-0.5em}
\noindent Without that boundary, a later user cannot tell when a research question has drifted beyond what the tool measures \citep{jacobs}.

\begin{RecBox}{T3 \textperiodcentered\ Contextuality}
Accept who is speaking, who is targeted, and prior context as scoring input. \end{RecBox}
\vspace*{-0.5em}
\noindent Structured metadata makes counterspeech and reclaimed language separable from the attacks they resemble \citep{yu-etal-2022-hate}. Where it is missing, the tool should report its absence rather than score as though context were irrelevant.

\begin{RecBox}{T4 \textperiodcentered\ Uncertainty}
Make the score reflect how much people agree, not the model's confidence. \end{RecBox}
\vspace*{-0.5em}
\noindent Annotator disagreement is signal about a contested construct, not noise \citep{leonardelli2021agreeing, uma2021learning}. A score tied to that agreement makes a contested case visible, where one normalised to an artificial class balance does not.

\begin{RecBox}{T5 \textperiodcentered\ Multilinguality}
Validate and report performance separately for every supported language. \end{RecBox}
\vspace*{-0.5em}
\noindent A single headline figure hides the disparities that matter most to the communities least well served, so per-language results need disclosure wherever performance falls below a defined threshold \cite{tonneau-etal-2025-hateday}.

\paragraph{Existing classifiers do not meet these requirements} \textsc{detoxify}, the most used open alternative, trains on the three Jigsaw Kaggle challenges and so inherits Jigsaw's construct (T2). It covers seven languages against \textsc{perspective}'s eighteen (T5), and its own documentation warns that profanity is scored toxic ``regardless of the tone or the intent of the author'', with bias ``towards already vulnerable minority groups'' \citep{detoxify}. OpenAI's moderation endpoint reproduces what created the original dependence: free at the point of use, a fixed thirteen-category taxonomy, and a single \texttt{omni-moderation-latest} alias behind which the model may change, meeting neither T1 nor T2 \citep{openai_moderation}. Open-weight guard models such as \textsc{llama guard} and \textsc{shieldgemma} improve on T3 by supplying the construct as a policy prompt at inference, but they move the choice of definition to whoever writes that prompt, and prompts are versioned no more carefully than models were (T1). No available classifier satisfies T1 and T2 together.

\begin{RecBox}{G1 \textperiodcentered\ Independence}
Keep the tool open-source, vendor-independent, and separate from what it evaluates. \end{RecBox}
\vspace*{-0.5em}
\noindent The shutdown is the argument. Independence is not a preference for open tools but rather is what lets an instrument outlive the company that built it, and the community repair it when it breaks.

\begin{RecBox}{G2 \textperiodcentered\ Transparency}
Open model weights, training data, and annotation schemata to external audit. \end{RecBox}
\vspace*{-0.5em}
\noindent Annotator demographics and beliefs shape what is labelled toxic in contested cases \citep{sap2022, davani2023, kraft2025socialbiaspopularquestionanswering}: annotation is part of the instrument, and publishing it is what lets a disparity be traced rather than merely observed.

\begin{RecBox}{G3 \textperiodcentered\ Legitimacy}
Define toxicity together with affected communities, and preserve disagreement between them. \end{RecBox}
\vspace*{-0.5em}
\noindent This should run through an open, documented process \citep{maronikolakis-etal-2022-listening}. Legitimacy does not require consensus. It requires that systematic disagreement, by annotator identity and target group, be documented and propagated rather than averaged away \citep{leonardelli2021agreeing, sap2022, davani2023}.

\begin{RecBox}{G4 \textperiodcentered\ Accountability}
Audit continuously, and publish both what the audits find and what was changed in response. \end{RecBox}
\vspace*{-0.5em}
\noindent Query-efficient methods now make external auditing tractable \citep{hartmann2026auditcanqueryefficientactive}, so what remains is a governance obstacle rather than a technical one. Accountability requires auditors independent of the maintainers, and a documented remediation for each finding.

\begin{RecBox}{G5 \textperiodcentered\ Sustainability}
Fund the infrastructure so that no single funder's exit can end it. \end{RecBox}
\vspace*{-0.5em}
\noindent \textsc{perspective api} was free at the point of use, which drove adoption and left no alternative when its funder exited. Free is not the same as sustainable: a successor needs funding spread across institutions, so that any one of them withdrawing slows maintenance rather than ending the service.

\paragraph{The governance templates exist} Working instances of these requirements are in place elsewhere. EleutherAI's \textit{lm-evaluation-harness}, the widely used open-source evaluation framework that advanced the community standard, implements T1 directly: each evaluation task carries a version that contributors increment on any breaking change, recorded with date, pull request, and description \citep{lmevalharness}. MLCommons is a template for G1, G4, and G5, being a 501(c)(6) non-profit whose AI Risk and Reliability working group spans industry, academia, and civil society, and which maintains \textsc{ailuminate} as a versioned public benchmark \citep{ailuminate}. Most directly, \textit{Dynabench} began inside Meta and was transferred to MLCommons in 2022 rather than being shut down \citep{kiela-etal-2021-dynabench}, the migration \textsc{perspective api} never made. None of these are toxicity tools, but each shows that the governance a successor needs is already running somewhere.

\paragraph{Two constraints and a first step} What none of these templates supplies is the measurement layer itself: versioned, per-language classifiers, and multiple construct-specific instruments under shared governance rather than one general toxicity tool. Two constraints bind its construction. The first is \textbf{privacy}. T3 asks a successor to score with sender, target, and conversation history, and G2 asks it to publish annotation schemata and annotator demographics, so both enlarge the personal data held about authors, targets, and annotators, on populations least able to absorb the risk. Our own release resolves a narrow version of this by publishing record identifiers and scores rather than the texts, and a successor faces the same choice at every layer. The second is \textbf{funding}. G5 asks for funding spread across institutions, but no durable model exists. Grant cycles are shorter than the maintenance horizon of infrastructure, and an instrument that must be revalidated as language changes cannot be funded as a one-off project. Both constraints are real, but the shutdown shows the cost of leaving them unsolved. As a first step, we call on ACL, ACM, and ICA to jointly establish a permanent working group covering versioned model oversight, annotation schema governance, and benchmark archiving, with a defined funding mechanism rather than merely institutional endorsement. Such a body could also take custody of what a provider retires, as MLCommons did for \textit{Dynabench}, while \textsc{perspective} shows what is lost when none does. 

\paragraph{A request to Jigsaw} With that precedent in mind, we ask Google Jigsaw to release the per-language model versions, annotation guidelines, and remaining score distributions. Doing this at deprecation would cost very little, but if not, a decade of toxicity research will rely on a tool that can no longer be reviewed. Jigsaw could also support a successor with transitional funding or by handing the project to a group that can maintain it, rather than issuing a sunset notice alone.

\section{Concluding Remarks} 
\label{sec:concluding}
The closure of \textsc{perspective api} is an opportunity for researchers to be honest about how the field came to depend on a tool it did not control. The community adopted a corporate tool, incorporated it into benchmark suites, declared it a standard measure, and continued using it after its validity problems were known. Why no alternative was built is as much a question about our epistemic practices as our technical ones.
This responsibility is not the community's alone. Google Jigsaw built \textsc{perspective api}, encouraged its use in research, and is now withdrawing it. It should release the model weights, annotation schemata, and annotation documentation that made it function as infrastructure, where ownership permits. A decade of publicly consequential work should not disappear at shutdown.
With this call for reflexivity and accountability, we join the active debate in critical research on dataset creation and use \citep{paullada2021data, sambasivan2021everyone}, while extending it to the measurement infrastructure that relies on and shapes those datasets. Past efforts at shared moderation infrastructure have not stalled for want of technical means but because such work counts as service labour. The community needs to recognise the scientific value of building measurement infrastructure, and to treat it as a form of care. Annotation schemas, inter-rater reliability scores, dataset construction decisions, and calibration evaluations deserve the same methodological rigour and visibility as model performance results. Funding bodies, programme committees, and tenure committees all participate in that judgement, and the community can change how such work is recognised and rewarded. Whether the field builds what replaces \textsc{perspective api} depends less on what is possible than on what it counts as research.
 \newpage

 \section*{Limitations}

\paragraph{No empirical evaluation} This is a position paper laying out evaluation criteria, so we do not empirically evaluate successor infrastructure against the requirements we propose (T1--T5, G1--G5). Building and testing systems against these criteria is the natural follow-up to this scope. Relatedly, our claims about path dependence and the self-reinforcing loop (\S\ref{sec:struct_circ}) rest on convergent qualitative evidence and the bibliographic search. Quantifying the drift and threshold sensitivity across the archived snapshot is a natural empirical extension. However, with the closure of the system, the focus should lie on learning from these issues.

\paragraph{Visibility bias} Our critique of measurement modelling focuses on the well-known drawbacks of \textsc{perspective api} and may not fully reflect situations where the API worked well and met its goals. Our critique is also shaped by what is visible. \textsc{perspective api} published model cards, architecture papers, training corpora, and its own bias evaluations, so its failures are unusually well documented. No comparable evidence base exists for its closed competitors, and the density of criticism here should not be read as evidence that it was worse than tools nobody can inspect.

\paragraph{No costing} We name funding as a binding constraint on the infrastructure we propose (\S\ref{sec:infra}) but do not price it. Compute, annotation, and long-term maintenance depend on scope, language coverage, and the institutional host, and we are not positioned to estimate them credibly from outside the bodies that would carry them. A costed proposal, developed with those bodies, is a necessary next step.

\paragraph{Partial stakeholder coverage} The requirements we propose reflect the main priorities of the research community and may not include all stakeholder views, especially those of practitioners, affected communities, or people from non-Western and non-English-speaking backgrounds.

\section*{Ethical Considerations}
This study does not involve human subjects. We did not collect data from participants or annotate any human-subject data. All labelling was done by our research team using published papers. The resource we share is a snapshot of Perspective API scores for 77 publicly available toxicity and hate-speech datasets. We only release record identifiers, scores, and query metadata. The original texts are not included and remain available from each dataset’s original release under its own license. Some source datasets are already distributed as identifiers or have access restrictions set by their authors. By releasing only identifiers, we respect these conditions. However, this means that to reconstruct the scored texts, users must retrieve the source datasets, and some records from platform-based corpora may no longer be accessible. The scores are released under a CC BY 4.0 license, and each source dataset keeps its own license and citation. The source datasets were designed to include offensive and distressing language. By sharing identifiers instead of texts, we reduce but do not completely prevent the spread of this content, since identifiers may still link to material hosted elsewhere. This archive is meant for evaluating hate-speech detection systems and studying measurement infrastructure. It should not be used to train models to generate hateful content.

This paper critiques a measurement tool that over-moderates identity-related and reclaimed language and under-moderates implicit hate. These failures fall unevenly on marginalised communities. We document these failures to support the development of more equitable measurement infrastructure, not to prescribe in advance what it should look like. The requirements in \S\ref{sec:infra} are offered in that spirit, since failures in hate-speech moderation can have serious consequences for the communities such moderation is meant to protect. As noted in our limitations, our account focuses on documented shortcomings and may under-represent contexts in which \textsc{perspective api} worked well. We also recognise that the requirements we propose reflect the priorities of our own research communities. As we argue under G3, legitimate successor infrastructure must be defined together with the communities most affected by toxicity measurement, and the requirements themselves should be validated against those communities' priorities.
\section*{LLM-based Tools}
We used writing assistance and LLM-based tools in a limited way during paper preparation and implementation. GitHub Copilot was used for code completion in the accompanying data release, Grammarly was used to suggest alternative phrasings, and Claude (Anthropic) was used for LaTeX formatting, copy-editing, and drafting suggestions on the abstract and Sections 2 to 5. All design decisions and substantive writing were carried out by the authors, who reviewed and revised every AI-assisted edit.

\section*{Acknowledgments}
This work was supported by the Weizenbaum Institute (grant number 16DII141), funded by the German Federal Ministry of Research, Technology, and Space (BMFTR) and the State of Berlin. Here, D.H., A.K., L.S., and J.B. specifically acknowledge the support of the Weizenbaum Institute in context of the Short Project \textit{Evaluating GenAI Evaluations}. This work also received funding from the European Union (EU) - NextGenerationEU and the Flemish Government under the ``Onderzoeksprogramma Artificiële Intelligentie (AI) Vlaanderen'' programme.

\bibliography{custom}
\newpage
\appendix
\section{Requirements Overview Table}
\label{app:requirements}
In Table \ref{tab:requirements}, we present the complete list of ten requirements, each linked to the specific failure in \S\ref{sec:critique} and \S\ref{sec:struct_circ} that inspired it. This mapping is meant to describe our process, not to serve as a final validation. It shows how we developed the requirements based on one tool's documented failures. The research community will need to test and refine these requirements as they apply them to new systems.

\begin{table*}[t!]
\centering
\small
\renewcommand{\arraystretch}{1.1}
\begin{tabular}{p{0.4cm}p{2.5cm}p{8.5cm}p{2.8cm}}
\toprule
\textbf{} & \textbf{Requirement} & \textbf{Recommendation} & \textbf{Addressed failure} \\
\midrule
T1 & Reproducibility & All metadata components of evaluations should be documented, i.e., metadata on the model, benchmark dataset, and evaluation metric \citep{biderman2024lessons}, in a community-agreed schema, and each release should be pinned to a citable version \citep{batzner2026everyevalever}. & Irreproducible, incompatible results \\

T2 & Validity & Evaluation tools should explicitly document which constructs, conditions, and input formats they measure, including limitations such as threshold selection and out-of-scope queries \citep{blodgett2021, jacobs, wallach2024}. & Construct underspecification \\

T3 & Contextuality & Scoring should support structured metadata input, namely who is speaking, who is targeted, and prior context, to enable classification of counterspeech and reclaimed language \citep{yu-etal-2022-hate}. & 
Misclassification of identity-laden content \\

T4 & Uncertainty & Evaluation tools should return a score that reflects how much annotators agree rather than model confidence, as that disagreement reflects contested constructs \citep{leonardelli2021agreeing, uma2021learning}. & Score unrelated to annotator agreement \\

T5 & Multilinguality & Performance must be validated and documented per language, with explicit disclosure where it falls below a defined threshold \citep{nogara2025, tonneau-etal-2025-hateday}. & Uneven cross-lingual performance \\

\midrule
G1 & Independence & The infrastructure should be open-source and institutionally independent of any commercial entity, so no single organisation controls annotation schema, model weights, or the definition of toxicity, as \textsc{perspective api} demonstrated \cite{rieder2021, pozzobon-etal-2023-challenges}. No instrument should both label training data and grade the models built from it. & Single-vendor dependence, circular evaluation \\

G2 & Transparency & Model weights, training data, and annotation schemata should be open and auditable, including annotator demographics, which influence group-level disparities \citep{sap2022, davani2023, kraft2025socialbiaspopularquestionanswering}. & Opacity of training and annotation \\

G3 & Legitimacy & Defining toxicity, selecting communities, and collecting training data must involve affected communities through an open, documented process \citep{maronikolakis-etal-2022-listening}. Legitimacy does not require consensus. It requires that systematic disagreement, by annotator identity and target group, be documented and propagated rather than averaged away \citep{leonardelli2021agreeing, sap2022, davani2023}. & Contested disagreement averaged away \\

G4 & Accountability & Auditing with published results and documented 
remediation across demographic groups, languages, and content types 
should be a continuous, open process run by auditors independent of the maintainers \citep{10.1145/3580494, hartmann2026auditcanqueryefficientactive}. & Undetected systematic errors \\

G5 & Sustainability & A long-term funding model must be put in place that does not reproduce the single-funder dependence that made \textsc{perspective api}'s shutdown so disruptive \citep{rieder2021}. & Funding dependence \\
\bottomrule
\end{tabular}
\caption{Ten requirements and recommendations for successor toxicity measurement infrastructure, derived from the structural failures documented in \S\ref{sec:critique}.}
\label{tab:requirements}
\end{table*}
\section{Bibliographic Review of Papers that used \textsc{perspective}}
\label{app:bibliography}

\subsection{Method}
We conducted a structured bibliographic search across ACL Anthology, Web of Science, and OpenAlex, identifying 768 records (383, 14, and 371 respectively). After removing 216 cross-database duplicates and 51 records excluded by rule-based screening (year $<$ 2017, off-topic titles, or keyword-collision false positives), 501 records remained for title/abstract screening, of which full text was sought for 454 and retrieved for 397.

For each retrieved paper, we extracted every mention of ``Perspective API'', ``Jigsaw'', or a bare ``perspective'' occurring near a toxicity-related term, taking a $\pm$400-character context window around each match. \textbf{All papers were labelled by hand by the first author}, using these extracted snippets to decide (a) whether the paper uses \textsc{perspective} in its own research and (b) its usage type -- baseline model, CSS measurement, pre-training filtering, LLM evaluation, or meta-critique -- consulting the full paper directly whenever the snippets left this ambiguous.

An LLM (\texttt{gpt-5-mini}), given the same snippets, was used only to cross-check the human labels, never to make the primary decision. Agreement between the LLM and the first author was 386/397 (97.2\%) on inclusion and 80.3\% on usage type. Every non-agreement was then reconsidered by hand against the full paper. Three usage-type labels were changed as a result; for all remaining cases the human label was retained.

Beyond the structured search, we separately traced the citations of every dataset we rescored (Appendix~\ref{app:snapshot}) back to its own originating paper. This citation-chaining step is not part of the systematic search above, but surfaced 12 further papers not caught by the original search terms, itself a small illustration of how fragmented and hard to fully enumerate this literature already is.

In total we identify 241 papers using \textsc{perspective} in their own research (223 from the four categories discussed in the main text, plus 18 meta-critique papers, for which \textsc{perspective} is the object rather than the instrument of study). Table~\ref{tab:usage-overview} gives the full breakdown. All paper categorisations can be found in the accompanying snapshot dataset \cite{hartmann2026snapshot}.

\begin{table}[h]
\centering
\begin{tabular}{lr}
\toprule
Usage type & $n$ \\
\midrule
Baseline model & 83 \\
LLM evaluation & 61 \\
CSS measurement & 50 \\
Pre-training filtering & 29 \\
Meta-critique & 18 \\
\midrule
Total & 241 \\
\bottomrule
\end{tabular}
\caption{Papers using \textsc{perspective}, by usage type (human-labelled).}
\label{tab:usage-overview}
\end{table}

\subsection{PRISMA Flow}
\begin{figure*}[h!]
    \centering
    \includegraphics[width=0.5\linewidth]{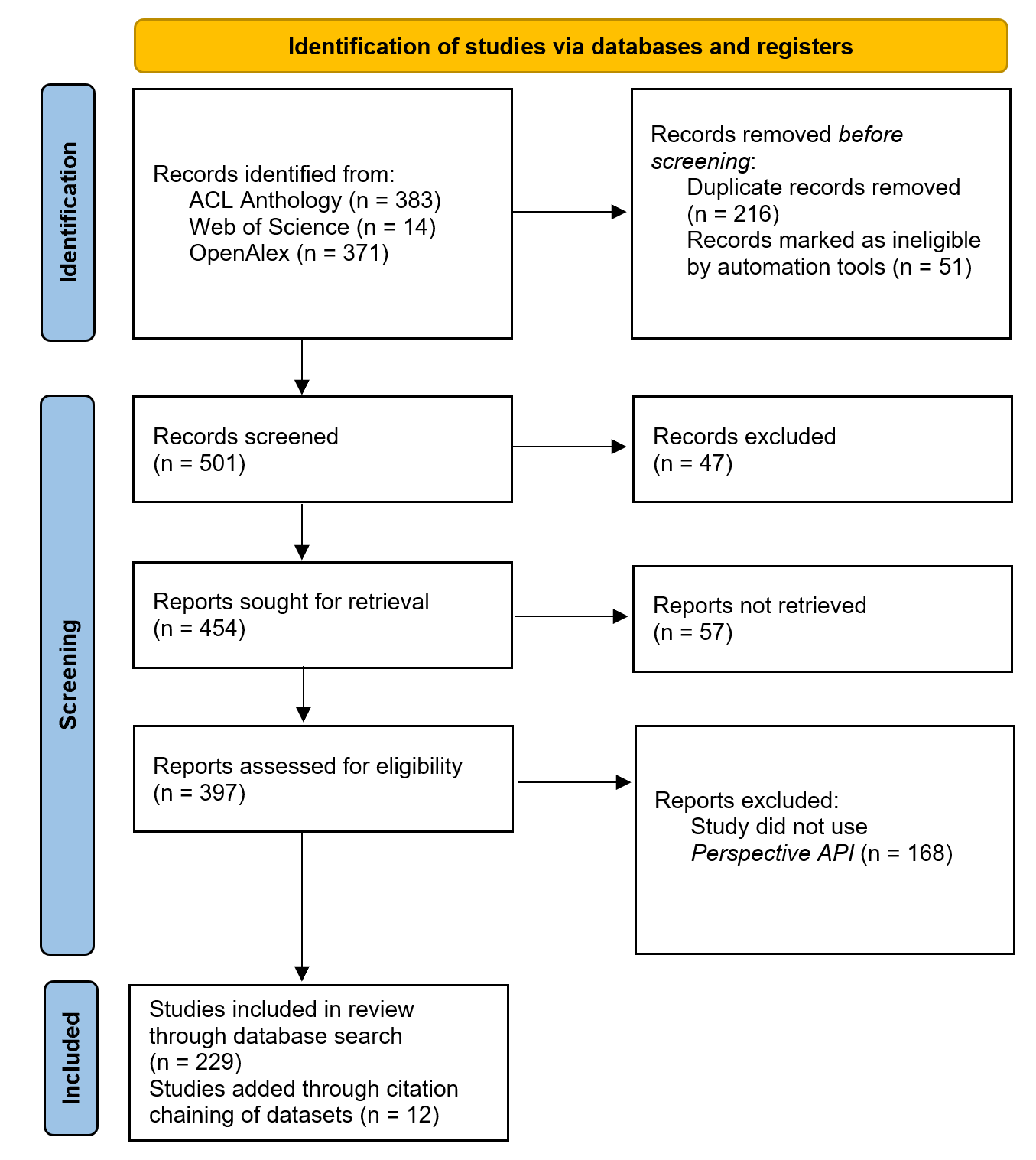}
    \caption{PRISMA flow diagram for the bibliographic search. Of 768 records identified across ACL Anthology, Web of Science, and OpenAlex, 229 were confirmed to use \textsc{perspective} in their own research. A further 12 were added through citation chaining from the datasets we rescored (Appendix~\ref{app:snapshot}), giving 241 in total. Screening was done by the first author on verbatim full-text snippets of \textsc{perspective api} usage, with LLM-assisted screening used as a check on agreement (97.2\% agreement on inclusion, 80.3\% on usage type). After the check, all non-agreements were reconsidered and 3 papers were changed; for the rest, the human label was used.}
    \label{fig:prisma}
\end{figure*}

Figure~\ref{fig:prisma} reports identification through inclusion following PRISMA~2020: 768 records identified, 267 removed before screening (216 duplicates and 51 excluded by automation), 501 screened, 47 excluded at title/abstract, 454 sought for retrieval, 57 not retrieved, 397 assessed with full text, 168 excluded as not using \textsc{perspective}, and 229 included by human review. A second, non-systematic identification stream, citation-chaining from the archived dataset snapshot, added 12 further records outside this flow, as described above, giving \textbf{241 papers} that used \textsc{perspective api}. Most excluded papers used Jigsaw's Kaggle datasets but did not use \textsc{perspective} for scoring.

\section{The Archived Snapshot}
\label{app:snapshot}
\subsection{Method}
Datasets were selected by first identifying every dataset cited in this paper's corpus with a publicly available release and scoring it. We then traced outward in three further ways. (i) Known, curated Hugging Face releases of hate-speech and toxicity datasets, including several used to construct multilingual superset benchmarks. (ii) A scan of every GitHub, OSF, Zenodo, and figshare repository already retrieved as a code artifact of a paper in our corpus, using both filename-based and content-based text-column detection, the latter to catch datasets whose text field is not conventionally named. (iii) A targeted search, per paper without an identified code artifact, for a corresponding public release. Every dataset was manually checked against its cited source paper before inclusion. Ambiguous or unverifiable matches were excluded rather than guessed.
\subsection{Overview}
In total, the snapshot covers \textbf{77 datasets and major benchmarks, totalling 6,162,826 records, of which 5,897,618 (95.7\%) were scored successfully}. Failures are concentrated in languages \textsc{perspective} does not support at all (e.g.\ Turkish, Finnish), which we report explicitly per dataset rather than omitting.

Each dataset was scored via \textsc{perspective}'s \texttt{v1alpha1} \texttt{comments:analyze} endpoint for all six production attributes (toxicity, severe toxicity, identity attack, insult, profanity, threat), with \texttt{doNotStore} set on every request. An explicit language hint was sent only for \textsc{perspective}'s documented production-supported languages. Otherwise the request relied on \textsc{perspective}'s own language auto-detection. Scoring took place between 2026-05-22 and 2026-08-28. \textsc{perspective} exposes no public model-version identifier, so we record a per-record query timestamp as the closest available version pin. We release, per dataset: a stable dataset and record identifier (sufficient to merge back onto the original release), the six attribute scores, the declared and \textsc{perspective}-detected language, success/failure status and error reason where applicable, and the query timestamp, not the underlying text, which remains available from each dataset's original release under its own license. Each dataset ships with a short README documenting its source, license pointer, and merge key. \textsc{perspective} scores are released under CC-BY 4.0. Full per-paper and per-dataset provenance, which paper each dataset is linked to, its citation, and licensing, is provided as a table in the accompanying repository:  \url{https://huggingface.co/datasets/dawiet/ByeByePerspectiveAPI-snapshot }. An overview of datasets can be seen in Table \ref{tab:snapshot}.

\onecolumn

\begin{longtable}{p{0.30\textwidth}p{0.19\textwidth}p{0.06\textwidth}p{0.24\textwidth}r}
\toprule
Dataset & Paper & Lang. & Category & $n$ rows \\
\midrule
\endfirsthead
\toprule
Dataset & Paper & Lang. & Category & $n$ rows \\
\midrule
\endhead
\bottomrule
\endfoot
\bottomrule
\caption{Archived snapshot: rescored datasets, their corresponding paper, dominant language(s), usage category, and row count. Rows labelled \emph{Hate Speech Superset} are language- and geography-specific subsets of the superset released by \citet{tonneau-etal-2024-languages}. Rows marked \emph{(mirror)} are separate public releases of the same underlying data, scored and released independently; their records are included in the totals in \S\ref{app:snapshot}.}
\label{tab:snapshot}
\endlastfoot
CivilComments & \citet{borkan2019}; \citet{dixon2018} & en & Baseline model & 1,968,756 \\
PolygloToxicityPrompts & \citet{jain2024ptp} & multi & LLM evaluation & 847,420 \\
HateDay & \citet{tonneau-etal-2025-hateday} & multi & Baseline model & 501,348 \\
Hate Speech Superset (Arabic) & \citet{tonneau-etal-2024-languages} & ar & Baseline model & 447,561 \\
DATG & \citet{liang2024datg} & en & LLM evaluation & 372,275 \\
RealToxicityPrompts & \citet{gehman2020} & en & LLM evaluation & 198,231 \\
Toraman et al. Hate Speech & \citet{toraman2022} & tr, en & Baseline model & 197,849 \\
Pile Toxicity (balanced) & \citet{korbak2023pretraining} & en & Baseline model & 196,759 \\
XDetox & \citet{lee2024xdetox} & en & LLM evaluation & 178,024 \\
Wikipedia Detox & \citet{wulczyn2017} & en & Baseline model & 159,117 \\ 
EFI & \citet{jiang2025probefree} & en & LLM evaluation & 90,038 \\
Wiki Madlibs & \citet{dixon2018} & en & LLM evaluation & 89,483 \\ 
Unintended Harms & \citet{choi2025unintendedharms} & en & LLM evaluation & 59,814 \\
Alpaca & \citet{taori2023alpaca} & en & LLM evaluation & 51,004 \\
Hate Speech Superset (German) & \citet{tonneau-etal-2024-languages} & de & Baseline model & 50,254 \\
TuPy-E & \citet{oliveira2023tupye} & pt & Baseline model & 43,210 \\
Hate Speech Superset (Portuguese) & \citet{tonneau-etal-2024-languages} & pt & Baseline model & 43,201 \\
Hate Speech Superset (Turkish) & \citet{tonneau-etal-2024-languages} & tr & Baseline model & 41,315 \\
Hatevolution & \citet{dibonaventura2025} & en & Baseline model & 41,133 \\
Measuring Hate Speech & \citet{kennedy2020interval}; \citet{sachdeva2022mhs} & en & Baseline model & 39,564 \\
Lost in Moderation & \citet{hartmann2025} & en & Baseline model & 35,491 \\
WritingPrompts & \citet{fan2018writingprompts} & en & LLM evaluation & 30,746 \\
Hate Speech Superset (Spanish) & \citet{tonneau-etal-2024-languages} & es & Baseline model & 29,646 \\
Davidson et al. HSOL & \citet{davidson2017} & en & Baseline model & 24,782 \\
PclGPT (Pcl-PT/SFT) & \citet{wang2024pclgpt} & en & Pre-training filter & 24,052 \\ 
Soteria & \citet{banerjee2025soteria} & ar & Pre-training filter & 22,793 \\ 
CantTalkAboutThis & \citet{rebedea2024canttalk} & en & Baseline model & 22,598 \\
Aligned Probing & \citet{waldis2026alignedprobing} & en & LLM evaluation & 21,969 \\
ToLD-Br & \citet{leite2020toldbr} & pt & Baseline model & 20,813 \\
EDOS & \citet{kirk2023edos} & en & Pre-training filter & 20,000 \\
HatEval & \citet{basile2019hateval} & en, es & Baseline model & 19,341 \\
HateXplain & \citet{mathew2021hatexplain} & en & Baseline model & 19,192 \\
Hate Speech Superset (French) & \citet{tonneau-etal-2024-languages} & fr & Baseline model & 17,879 \\
LM-Steer & \citet{han2024lmsteer} & en & LLM evaluation & 17,394 \\
Capybara & --- & en & Pre-training filter & 15,935 \\ 
Mix and Match & \citet{mireshghallah2022mixmatch} & en & LLM evaluation & 15,144 \\
Hate Speech Superset (Indonesian) & \citet{tonneau-etal-2024-languages} & id & Baseline model & 14,184 \\
Hate Speech Superset (India) & \citet{tonneau-etal-2024-languages} & multi & Baseline model & 14,095 \\
Lukito et al. Connective Data & \citet{lukito2024connective} & en & CSS measurement & 12,151 \\ 
Ghosh Roy et al. Multilingual HS & \citet{ghoshroy2021multilingual} & en & Baseline model & 12,072 \\ 
CONAN & \citet{chung2019conan} & en & Baseline model & 11,641 \\
GAHD & \citet{goldzycher2024gahd} & de & Pre-training filter & 10,994 \\
GAHD (mirror) & \citet{goldzycher2024gahd} & de & Pre-training filter & 10,994 \\
ToxBART & \citet{yadav2024toxbart} & en & Baseline model & 10,423 \\
ToxicChat & \citet{lin-etal-2023-toxicchat} & en & Pre-training filter & 9,718 \\
Arango Monnar et al. (Chilean) & \citet{arango-monnar-etal-2022-resources} & es & Baseline model & 9,216 \\     
DiaSafety & \citet{sun2022diasafety} & en & Baseline model & 7,692 \\
HateBR & \citet{vargas2022hatebr} & pt & Baseline model & 7,000 \\
HateCheckHIn & \citet{das2022hatecheckhin} & hi & Baseline model & 5,884 \\
Fortuna et al. Portuguese HS & \citet{fortuna2019} & pt & Baseline model & 5,668 \\
TextDetox (Japanese) & \citet{dementieva2025detox} & ja & Baseline model & 4,908 \\
HatemojiBuild & \citet{kirk2022hatemoji} & en & Baseline model & 4,723 \\
HateCheck & \citet{rottger2021} & en & Baseline model & 3,728 \\
Quantifying Stereotypes & \citet{liu2024stereotypes} & en & Baseline model & 2,960 \\
YouNICon & \citet{liaw2023younicon} & en & CSS measurement & 2,528 \\
Reddit-BR Toxicity & \citet{lima2024redditbr} & pt & Baseline model & 2,500 \\
Instructive Debiasing & \citet{morabito2023debiasing} & en & LLM evaluation & 2,398 \\
Integrative Decoding & \citet{cheng2024integrative} & en & LLM evaluation & 2,271 \\
Suomi24 Toxicity & --- & fi & Baseline model & 2,236 \\ 
Conversations Gone Awry & \citet{zhang2018awry} & en & Pre-training filter & 2,169 \\
Perspective API Eval Set & \citet{lees2022} & en & Baseline model & 1,997 \\
AustroTox & \citet{pachinger2024austrotox} & de & Pre-training filter & 1,983 \\
AustroTox (mirror) & \citet{pachinger2024austrotox} & de & Pre-training filter & 1,983 \\
Constructive Deliberation & \citet{kambhatla2024deliberation} & en & Pre-training filter & 1,956 \\
RSA-Control & \citet{wang2024rsacontrol} & en & LLM evaluation & 1,507 \\
Chilean Protests (Spanish) & \citet{gonzalezbustamante2024} & es & Baseline model & 1,000 \\
Chilean Constitution (Spanish) & \citet{gonzalezbustamante2024} & es & Baseline model & 1,000 \\
ETHOS & \citet{mollas2022ethos} & en & Baseline model & 997 \\
Barber\'a Twitter Ideology & \citet{barbera2015} & en & Baseline model & 945 \\ 
SimulBench & \citet{jia2025simulbench} & en & CSS measurement & 753 \\ 
eMFD & \citet{hopp2021emfd} & en & Pre-training filter & 495 \\ 
Ross et al. Refugee Tweets & \citet{ross2016} & de & Baseline model & 469 \\
PPLM & \citet{dathathri2020pplm} & en & Pre-training filter & 440 \\
Text vs. Image Hate Detection & \citet{aggarwal2024textorimage} & en & Baseline model & 407 \\
Multilingual Blending & \citet{song2025blending} & multi & LLM evaluation & 360 \\
AP-Test & \citet{yang2026guardrail} & en & Baseline model & 150 \\
SimpleSafetyTests & \citet{vidgen2023sst} & en & LLM evaluation & 100 \\
\end{longtable}
\end{document}